\pdfoutput=1
\documentclass[11pt]{article}

\usepackage{emnlp2021}

\usepackage{times}
\usepackage{latexsym}

\usepackage[T1]{fontenc}
\usepackage[utf8]{inputenc}

\usepackage{microtype}

\usepackage{algorithm}
\usepackage{algorithmic}
\usepackage{enumerate}
\usepackage{CJKutf8}
\usepackage{url}
\usepackage{latexsym}
\usepackage{graphicx}
\usepackage{times}
\usepackage{latexsym}
\usepackage{amsmath}
\usepackage{amssymb}
\usepackage{array}
\usepackage{comment}
\usepackage{multirow}
\usepackage[english]{babel}
\usepackage{epstopdf}
\usepackage{times}
\usepackage{url}
\usepackage{latexsym}
\usepackage{graphicx}
\usepackage{amsmath}
\usepackage{amssymb}
\usepackage{amsmath}
\usepackage{mathrsfs}
\usepackage{booktabs}
\usepackage{mathtools}
\usepackage{pifont}
\usepackage[utf8]{inputenc}
\usepackage{algorithm}
\usepackage{algorithmic}
\usepackage{caption}
\usepackage{graphics}
\usepackage{color}
\usepackage{booktabs}

\usepackage{diagbox}
\usepackage{algorithm}
\usepackage{algorithmic}
\usepackage{enumerate}
\usepackage{CJKutf8}
\usepackage{url}
\usepackage{latexsym}
\usepackage{graphicx}
\usepackage{times}
\usepackage{latexsym}
\usepackage{amsmath}
\usepackage{array}
\usepackage{comment}
\usepackage{multirow}
\usepackage[english]{babel}
\usepackage{epstopdf}
\usepackage{times}
\usepackage{url}
\usepackage{latexsym}
\usepackage{graphicx}
\usepackage{amsmath}
\usepackage{amssymb}
\usepackage{amsmath}
\usepackage{mathrsfs}
\usepackage{booktabs}
\usepackage{mathtools}
\usepackage{pifont}
\usepackage[utf8]{inputenc}
\usepackage{algorithm}
\usepackage{algorithmic}
\usepackage{subfigure}
\usepackage{bbding}

\title{Multilingual Machine Translation Systems from Microsoft for \\ WMT21 Shared Task}
\newcommand\ourmethod{DeltaLM + Zcode}
\author{
 Jian Yang,
 Shuming Ma, 
 Haoyang Huang,
 Dongdong Zhang,
 Li Dong,
 Shaohan Huang,\\
 \textbf{Alexandre Muzio,}
 \textbf{Saksham Singhal,}
 \textbf{Hany Hassan Awadalla,}
 \textbf{Xia Song,}
 \textbf{Furu Wei}
 \\
 Microsoft Corporation\\
 \texttt{\{t-jianya,shumma,haohua,dozhang,lidong1,shaohanh\}@microsoft.com} \\
 \texttt{\{alferre,saksingh,hanyh,xiaso,fuwei\}@microsoft.com} }

\begin{document}
\maketitle
\begin{abstract}
% We describe Microsoft's multilingual systems for the WMT21 shared task on large-scale multilingual machine translation. We participate in all of the two Small Tracks and Large Track. Our systems are fine-tuned from DeltaLM\footnote{https://aka.ms/deltalm}, a generic pre-trained multilingual encoder-decoder model. Different from Small Tracks, Large Track is fully unconstrained, so we collect the data in 102 languages from different sources. We also explore some approaches (i.e. progressive learning and iterative back-translation) to further improve the performance. Our final submission ranks first across three tracks in terms of automatic evaluation metrics.
This report describes Microsoft's machine translation systems for the WMT21 shared task on large-scale multilingual machine translation. We participated in all three evaluation tracks including Large Track and two Small Tracks where the former one is unconstrained and the latter two are fully constrained. Our model submissions to the shared task were initialized with DeltaLM\footnote{\url{https://aka.ms/deltalm}}, a generic pre-trained multilingual encoder-decoder model, and fine-tuned correspondingly with the vast collected parallel data and allowed data sources according to track settings, together with applying progressive learning and iterative back-translation approaches to further improve the performance. Our final submissions ranked first on three tracks in terms of the automatic evaluation metric.
\end{abstract}

\section{Introduction}
Recently, multilingual neural machine translation has attracted lots of attention because it enables one model to translate between multiple languages \cite{ls_decoder,google_mnmt,mnmt_challenges,mnmt_survey,monolingual_adapter,ls_subnet}. To improve the performance of the multilingual translation models, there are various approaches on the training methods \cite{mmnmt,balancec_mnmt,negative_interference_mnmt}, the model structures \cite{one_to_many_mnmt,adaptive_sparse_transformer,share_or_not}, and the data augmentation \cite{distillation_mnmt,ctl_mnmt}. M2M~\cite{m2m} leverages the large-scale data mined from the web data and explore the strategies to scale the model size and train the model effectively. Meanwhile, the multilingual pre-trained language models have proven beneficial for the multilingual machine translation models. mBART~\cite{mbart} pre-trains a multilingual model with the multilingual denoising objective to improve the multilingual machine translation.

% Despite the success of these advanced technologies, most of their effects on the very large-scale data have not been verified. 
% M2M~\cite{m2m} leverages the large-scale data mined from the web data and explore the strategies to scale the model size and train the model effectively. However, it doesn't use the pre-trained model. mBART~\cite{mbart} pre-train a multilingual model with the multilingual denoising objective to improve the multilingual machine translation, but its effect when combining with other approaches has not been explored.

In this work, we explore the effects of different advanced approaches for multilingual machine translation models, especially on the large-scale dataset.
We first explore the way to leverage the pre-trained language models that have been trained with large-scale monolingual data.
We use the public available DeltaLM-Large checkpoint to initialize the model. DeltaLM~\cite{deltalm} is a multilingual pre-trained encoder-decoder model, which has been proven useful for multilingual machine translation. 

We further explore the training methods and the data augmentation to improve the model.
For efficient training, we apply progressive learning \cite{shallow_to_deep,progressive_learning_multitask,progressive_learning_noise} to our model that continue-trains a shallow model into a deep model. Specifically, we first train a model with 24 encoder layers, and then continue-train it by adding 12 layers on the top of the encoder. As for the data augmentation, we implement iterative back-translation \cite{iterative_bt,weight_iterative_bt} that back-translates the data for multiple rounds. Due to the limits of time and GPU memories of the shared task, we do not explore other approaches like mixture-of-experts (MOE) and model ensemble.

We participated in all three tracks including Large Track, Small Track \#1, and Small Track \#2. Our final submissions are fine-tuned from DeltaLM with the allowed data sources according to the track settings, followed by progressive learning and iterative back-translation. The submissions on three tracks all rank first in terms of the automatic evaluation metric.

\section{Data}

%%%%%%%%%%%%%%%%%%%%%%%%%%%%%%%%%%%%%%%%%%%%%%%%%%%%%%%%%%%
\begin{figure*}[t]
\begin{center}
	\includegraphics[width=0.9\textwidth]{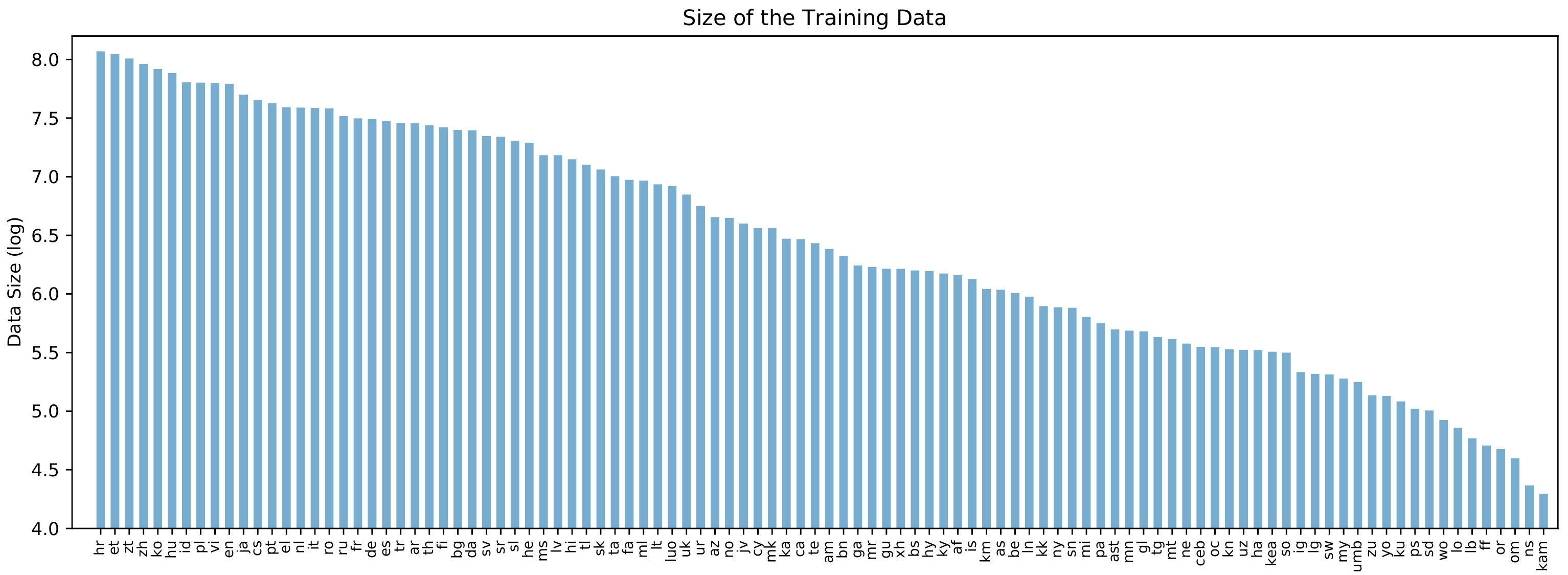}
	\caption{Dataset statistics of the bilingual data of the 102 languages. For better visualization, we apply the logarithmic function (base 10 logarithm) to the size of the training data. Each column denotes the data size of a language that was paired with the remaining 101 languages. For example, the first column denotes the number of bilingual sentence pairs that contain sentences from language hr.}
	\label{large_track_bitext_size}
%  	\vspace{-15pt}
\end{center}
\end{figure*}
%%%%%%%%%%%%%%%%%%%%%%%%%%%%%%%%%%%%%%%%%%%%%%%%%%%%%%%%%%%

\paragraph{Large Track} The monolingual and bilingual data are collected from multiple sources, including CCAligned \cite{ccalign}, CCMatrix \cite{ccmatrix}, OPUS-100 \cite{opus_100}, JW300 \cite{jw300}, Tatoeba \cite{tatoeba}, WMT2021 news track\footnote{\url{http://statmt.org/wmt21/translation-task.html}}, multilingual track data\footnote{\url{http://data.statmt.org/wmt21/multilingual-task/}}, and our in-house data. To improve the translation quality of non-English languages, we construct
dual-pseudo parallel data (or dual-pseudo data briefly) in which the source and target sides per each sentence pair are translated from the same monolingual English sentence respectively. The Wikipedia English monolingual sentences are translated to other 70 languages by leveraging various machine translation models including in-house MT models, M2M \cite{m2m}, the multilingual model of small tracks, and our intermediate multilingual MT model.

Finally, the training data was split into three parts: the bitext data (1.7B parallel sentences from 394 language pairs), the back-translation (1.4B parallel sentences from 45 language pairs), and the dual-pseudo data (8.7B parallel sentences of 70 languages from 4830 language pairs). Figure \ref{large_track_bitext_size} lists the statistics of the bilingual training data size of 102 languages. 

\paragraph{Small Track \#1} We use the constrained monolingual and bilingual data of 6 languages (Croatian, Hungarian, Estonian, Serbian, Macedonian, and English) provided by the shared task. According to the statistics, the bitext data contains 273M sentence pairs of all translation directions. Inspired by the previous work, we leverage the multilingual iterative back-translation method with one single multilingual model to generate parallel pseudo data.
%over a multilingual model to construct various parallel pseudo data.
For En$\to$X and X$\to$En directions, we generate the back-translation data of 390M sentence pairs. As for X$\to$Y directions, we generate the dual-pseudo data of 1.18B sentence pairs, where X and Y stand for any two non-English languages. 

%For additional data, we use back-translation data of 390M sentence pairs in En$\to$X and X$\to$En directions. As for X$\to$Y direction, dual-pseudo data contains 1180M sentence pairs, where X and Y stand for any two non-English languages. 

\paragraph{Small Track \#2} The monolingual and bilingual corpora of 6 languages (Javanese, Indonesian, Malay, Tagalog, Tamil, and English) provided by the shared task are used for the multilingual model training, containing 98M bilingual data, 256M generated back-translation data, and 860M generated dual-pseudo data.

\section{Large-scale Data Augmentation}
In this section, we introduce details about how to perform the iterative back-translation method~\cite{iterative_bt} to augment data.
% Due to lack of training data for many low-resource language pairs, we perform the iterative back-translation method~\cite{iterative_bt} to make data augmentation. 
We use different models for data augmentation according to different tracks.
For the small tracks, the multilingual models were trained over the constrained data sets to generate data. For the large track, we leverage the M2M model~\cite{m2m}, the intermediate multilingual MT models, and in-house MT models to generate different language pairs' data respectively, so as to play their respective advantages to enhance the data generation quality.
%and our to determine a more meaningful way, where the combination of different multilingual models provide the strong supplementary support for most low-resource languages. 
% in order to generate the high-quality back-translation (BT) data, 

In practice, both the monolingual and bilingual corpora are effectively utilized in three ways: 1) For the back-translation data of X$\to$En and En$\to$X directions, we used the best model to generated X data accordingly by back-translating monolingual English Wikipedia data;
2) For the dual-pseudo data of X$\to$Y directions, they are generated by back-translating the same English text to X and Y respectively. Alternatively, when the monolingual data of either X or Y is enough, we also directly perform back-translation between X and Y to obtain pseudo parallel data; 3) We try to augment existing bilingual corpora with the third language. Given the bilingual corpus ($X_{1}$, $Y_{1}$), we generate pseudo parallel corpus of ($X_{1}$, $Y_{2}$) and ($X_{2}$, $Y_{1}$) by back-translating $X_1$ to $X_2$ and $Y_1$ to $Y_2$, where $X_{2}$ and $Y_{2}$ are non-English languages.

%Besides back-translating the mono-lingual data, the available bilingual corpora can be used for enlarging the data. Then, the pseudo parallel corpus ($X_{1}$, $Y_{2}$) and ($X_{2}$, $Y_{1}$) are used as part of the augmented data. 

\section{Preprocessing}

\paragraph{Filtering} To enhance the model performance, we remove the noisy sentence pairs with the incorrect language identification or character encoding. More specifically, we remove the sentences longer than 1024 words and truncate the sentence to 512 tokens. We also construct three corpora after tokenization with different length ratio limitations, i.e. $\{1.5, 2.0, 2.5, 3.0\}$, between the source and the target sentence. Our multilingual model is first trained on the entire noisy data set and then continually tuned on cleaner data with descending length ratio, where the number of training directions is also gradually reduced by removing noisy language pairs. Therefore, we can progressively fine-tune the multilingual model in an efficient way (noisy corpora $\to$ clean corpora $\land$ numerous directions $\to$ selected directions $\land$ shallow encoder layers $\to$ deep encoder layers).
Besides, to clean the back-translation corpora, we remove the sentences containing unknown tokens (\texttt{[UNK]}). Regarding the language Sr (Serbian), those sentences comprised of Latin characters in training data were also discarded since we found that the validation sets use Cyrillic script for this language instead.

\paragraph{Tokenization} After data filtering, we use the SentencePiece \cite{spm} to tokenize all raw training, validation, and test data sets, where the SentencePiece model is consistent with the one used for DeltaLM \cite{deltalm}. We shuffled the whole training dataset before launching the training of multilingual models.
The input sentence is prefixed with the language tag to indicate the translation direction.

\section{Model and Training}
\subsection{DeltaLM}
We adopt the \texttt{DeltaLM\_large} architecture as the backbone model for all our experiments, which has 24 Transformer encoder layers and 12 interleaved decoder layers with an embedding size of 1024, a dropout of 0.1, the feed-forward network size of 4096, and 16 attention heads. We directly initialize our model with the public available DeltaLM large checkpoint\footnote{\url{https://aka.ms/deltalm}}. 

\subsection{Multilingual Fine-tuning}
The training data was split into the bitext corpora $D_{b}=\{D^{1}_{b},\dots,D^{u}_{b}\}$, the back-translation corpora $D_{bt}=\{D^{1}_{bt},\dots,D^{v}_{bt}\}$, and the dual-pseudo corpora $D_{dp}=\{D^{1}_{dp},\dots,D^{w}_{dp}\}$, where $u,v,w$ represent the number of the corpora of different translation directions.
The multilingual model with parameters $\Theta$ is jointly trained over the corpora to optimize the combined objective as below: 
\begin{SmallEquation}
\begin{align}
\begin{split}
    \mathcal{L}_{MT} =&-\lambda_{1}\sum_{i=1}^{u} \mathbb{E}_{x,y \in D_{b}^{i}} \left[ -\log P(y|x;\Theta) \right] \\
                    &-\lambda_{2}\sum_{i=1}^{v} \mathbb{E}_{x,y \in D_{bt}^{i}} \left[ -\log P(y|x;\Theta) \right] \\
                    &-\lambda_{3}\sum_{i=1}^{w} \mathbb{E}_{x,y \in D_{dp}^{i}} \left[ -\log P(y|x;\Theta) \right]
    \label{objective}
\end{split}
\end{align}
\end{SmallEquation}where $x,y$ denote the sentence pair in the bilingual corpus. $\mathcal{L}_{MT}$ is the combined translation objective of the multilingual model. $\lambda_{1}, \lambda_{2}$, $\lambda_{3}$ ($\lambda_1+\lambda_2+\lambda_3=1.0$) are used to balance the training objectives of the bitext corpora, the back-translation corpora, and the dual-pseudo corpora. In this work, we first set $\lambda_{1}=0.33, \lambda_{2}=0.33, \lambda_{3}=0.33$ and then reset $\lambda_{1}=0.6, \lambda_{2}=0.2, \lambda_{3}=0.2$ to focus more on the bitext corpora avoiding the noise introduced by pseudo data.

We follow the dynamic temperature-based data-sampling strategy~\cite{m2m,zcode} to ease the underrepresentation of low-resource languages. The probability of picking a language is proportional to its number of sentences $D_l$, i.e., $p_l=\frac{D_l}{\sum_i{D}_i}$. We set the temperature $T=5$ to rescale and control the distribution $p_l^\frac{1}{T}$. It can balance the samples between the high-resource languages and the low-resource languages.

\subsection{Progressive Learning}

We implement the progressive training method to train the model from shallow to deep \cite{shallow_to_deep}. The training process can be divided into two stages. In the first stage, the pre-trained DeltaLM model with 24 encoder layers and 12 decoder layers is directly adopted to initialize the multilingual translation model with the same architecture. The shallow translation model with 24 encoder layers and 12 decoder layers is fine-tuned on all available multilingual corpora. In the second stage, we increase the depth of the encoder from 24 layers to 36 layers, where the bottom 24 layers of the encoder are initialized with the shallow model's encoder and the top 12 layers are randomly initialized. Then we perform continue training. The deeper encoders enlarge the model's capacity, but no much extra decoding cost is introduced.

%2) We further increase the depth of the encoder to the 36 layers initialized from the shallow translation model with the 24 encoder layers and 12 decoder layers. The additional 12 layers of the encoder are randomly initialized. The deeper encoders enlarge the model's capacity without much extra decoding time cost.
% \section{Experiments}
%%%%%%%%%%%%%%%%%%%%%%%%%%%%%%%%%%%%%%%%%%%%%%%%%%%%%%%%%%%%%%%%%%%%%%%%%%%%
\begin{table*}[t]
\centering
\resizebox{0.8\textwidth}{!}{
\begin{tabular}{l|ccc|ccc|c}
\toprule
                  & \#Languages & \#Params & \#Layers & Avg$_{X \to En}$ & Avg$_{En \to Y}$ &  Avg$_{X \to Y}$ &Avg$_{all}$  \\
\midrule
\multirow{2}{*}{M2M \cite{m2m}} &102 &175M &6/6   &15.43 &12.02 &5.85   &6.00   \\
                                &102 &615M &12/12 &20.03 &16.21 &7.66   &7.86   \\ \midrule
\multirow{3}{*}{\ourmethod{} (Direct)}  &102 &711M &24/6  &30.39 &23.52 &11.21  &11.52   \\
     &102 &862M &24/12 &33.09 &27.21 &13.56  &13.89   \\
      &102 &1013M &36/12 &33.35 &27.39 &14.34 &14.65   \\ \midrule
\multirow{3}{*}{\ourmethod{} (Pivot)}  &102 &711M  &24/6  &31.32      &24.04   &14.74&14.99        \\
      &102 &862M  &24/12 &33.09 &27.21  &17.20  &17.45        \\
      &102 &1013M &36/12 &33.35      &27.39      &17.36      &17.62       \\ \midrule
\multirow{3}{*}{\bf \ourmethod{} (Hybrid)}    &102 &711M  &24/6  &31.32&24.04 &14.76& 15.01        \\
      &102 &862M  &24/12 &33.09      &27.21      &17.27      &17.52        \\
      &102 &1013M &36/12 &\bf 33.35      &\bf 27.39      &\bf 17.44      &\bf 17.70       \\
\bottomrule
\end{tabular}}
\caption{Evaluation results of Large Track for M2M and our method of 102 languages on the devtest of the FLORES-101 benchmark. Avg$_{X \to En}$ denotes the average score of directions between other languages to English. Avg$_{X \to En}$ denotes the average score of directions between English and other languages. Avg$_{X \to Y}$ denotes the average score of directions between non-English languages to other non-English languages. Avg$_{all}$ denotes the average result of all translation directions.}
\label{table:large_track}
\end{table*}
%%%%%%%%%%%%%%%%%%%%%%%%%%%%%%%%%%%%%%%%%%%%%%%%%%%%%%%%%%%%%%%%%%%%%%%%%%%%

%%%%%%%%%%%%%%%%%%%%%%%%%%%%%%%%%%%%%%%%%%%%%%%%%%%%%%%%%%%%%%%%%%%%%%%%%%%%
\begin{table*}[t]
\centering
\resizebox{0.8\textwidth}{!}{
\begin{tabular}{l|ccc|ccc|c}
\toprule
                 & \#Languages & \#Params & \#Layers & Avg$_{X \to En}$ & Avg$_{En \to Y}$ &  Avg$_{X \to Y}$ &Avg$_{all}$  \\
\midrule
\multirow{2}{*}{M2M \cite{m2m}}       &102 &175M &6/6   &24.60 &20.83 &20.80 &21.44   \\
                                      &102 &615M &12/12 &31.58 &29.62 &26.66 &27.98   \\ \midrule
\multirow{2}{*}{\bf \ourmethod{} (Direct)} &6   &862M   &24/12 &43.78 &41.02 &34.38 &37.06   \\
                                      &6   &1013M  &36/12 & \bf 44.34 &\bf 41.32 &\bf 34.68 &\bf 37.39   \\
\bottomrule
\end{tabular}}
\caption{Evaluation results of Small Track \#1 for M2M and our method of 6 languages (Croatian, Hungarian, Estonian, Serbian, Macedonian, English) on the devtest of the FLORES-101 benchmark. \textbf{\ourmethod{} (Direct)} denotes the strategy that we choose the direct translation for all translation directions, where the target language symbol is prefixed to the input sentence to indicate the translation direction. Our mutilingual translation model is only trained on the constrained corpora of 6 languages provided by the shared task.}
\label{table:small_track1}
% \vspace{-10pt}
\end{table*}
%%%%%%%%%%%%%%%%%%%%%%%%%%%%%%%%%%%%%%%%%%%%%%%%%%%%%%%%%%%%%%%%%%%%%%%%%%%%

%%%%%%%%%%%%%%%%%%%%%%%%%%%%%%%%%%%%%%%%%%%%%%%%%%%%%%%%%%%%%%%%%%%%%%%%%%%%
\begin{table*}[t]
\centering
\resizebox{0.8\textwidth}{!}{
\begin{tabular}{l|ccc|ccc|c}
\toprule
                 & \#Languages & \#Params & \#Layers & Avg$_{X \to En}$ & Avg$_{En \to Y}$ &  Avg$_{X \to Y}$ &Avg$_{all}$  \\
\midrule
\multirow{2}{*}{M2M \cite{m2m}}                 &102 &175M &6/6    &18.95 &15.16 &9.49  &12.01   \\
               &102 &615M &12/12  &24.67 &19.14 &12.11 &15.38   \\ \midrule
\multirow{2}{*}{\ourmethod{} (Direct)}     &6 &862M   &24/12  &43.12 &39.78 &28.69 &32.94   \\
                           &6 &1013M  &36/12  &43.56 &39.04 &28.60 &32.83   \\ \midrule
\multirow{2}{*}{\ourmethod{} (Pivot)}     &6 &862M   &24/12  & 43.12 & 39.78 & 29.02 & 33.17\\
                           &6 &1013M  &36/12   & 43.56 & 39.04 & 28.63 & 32.85\\\midrule
\multirow{2}{*}{\bf\ourmethod{} (Hybrid)}    &6 &862M  &24/12   &\bf 43.12  &\bf 39.78  &\bf 29.38  &\bf 33.40    \\
                           &6 &1013M  &36/12  & 43.56 & 39.04 & 28.99 & 33.09\\
\bottomrule
\end{tabular}}
\caption{Evaluation results of Small Track \#2 for M2M and our method of 6 languages (Javanese, Indonesian, Malay, Tagalog, Tamil, English) on the devtest of the FLORES-101 benchmark. \textbf{\ourmethod{} (Hybrid)} denotes the strategy that we choose the pivot-based translation (X$\to$En, En$\to$X) for low-resource X$\to$Y directions and direct translation for high-resource X$\to$Y directions.}
% \vspace{-10pt}
\label{table:small_track2}
\end{table*}
%%%%%%%%%%%%%%%%%%%%%%%%%%%%%%%%%%%%%%%%%%%%%%%%%%%%%%%%%%%%%%%%%%%%%%%%%%%%

\subsection{Training Details}

We train multilingual models with the Adam optimizer~\cite{adam} ($\beta_{1}=0.9$, $\beta_{2}=0.98$). The learning rate is set as 1e-4 with a warm-up step of $4,000$. The models are trained with the label smoothing with a ratio of 0.1. All experiments are conducted on 64 NVIDIA V100 or 32 A100 GPUs. The batch size is 1536 or 2048 tokens per GPU and the model is updated every 32 (for 64 V100 GPUs) or 64 (for 32 A100 GPUs) steps to simulate the large batch size.

\subsection{Decoding}

To enhance the performance of low-resource language pairs for X$\to$Y directions, we adopt the pivot-based translation method \cite{pivot_based_transfer_learning}. We use English as the pivot language and employ a unified model to perform the pivot-based translation. When the performance of X$\to$Y directions on the validation set is better than the pivot-based translation X$\to$En $\land$ En$\to$Y, we directly translate the language $X$ into $Y$. Otherwise, we translate them in the pivot way.
This approach is used for the submission to Large Track and Small Track \#2. As for Small Track \#1, we do not use the pivot-based translation.
% Therefore, the hybrid strategy including the direct translation and pivot-based translation is adopted for Large Track and Small Track \#2, where both tasks contain extremely unbalanced corpora of all translation directions. Considering the balanced distribution of the training data in all translation directions, the multilingual model only with the direct translation is submitted to Small Track \#1.

% \input{model}

\section{Evaluation Results}

Following the previous work~\cite{flores_101},  we use the dev and the devtest of the FLORES-101 benchmark as our validation set and test set respectively. During the inference, the beam search strategy is performed with a beam size of $4$ for the target sentence generation. We set the length penalty as $1.0$ by default. The last $N$ checkpoints ($N=\{1,5,10, 15, 20\}$) are averaged for evaluation and we select the best checkpoint based on the performance on the validation set. We report the SentencePiece-based BLEU using spBLEU\footnote{\url{https://github.com/ngoyal2707/sacrebleu.git}}.

% \subsection{Baselines}
% \paragraph{M2M} M2M is a strong many-to-many multilingual translation baseline that can translate directly between any pair of 100 languages. In this work, we compare our method with the M2M base\footnote{\url{https://dl.fbaipublicfiles.com/flores101/pre-trained_models/flores101_mm100_175M.tar.gz}} and large\footnote{\url{https://dl.fbaipublicfiles.com/flores101/pre-trained_models/flores101_mm100_615M.tar.gz}} models provided by the shared task. The two models support most languages of the WMT large-scale multilingual machine translation track.

\subsection{Large Track} 
\begin{figure*}[t]
\begin{center}
	\includegraphics[width=1.0\textwidth]{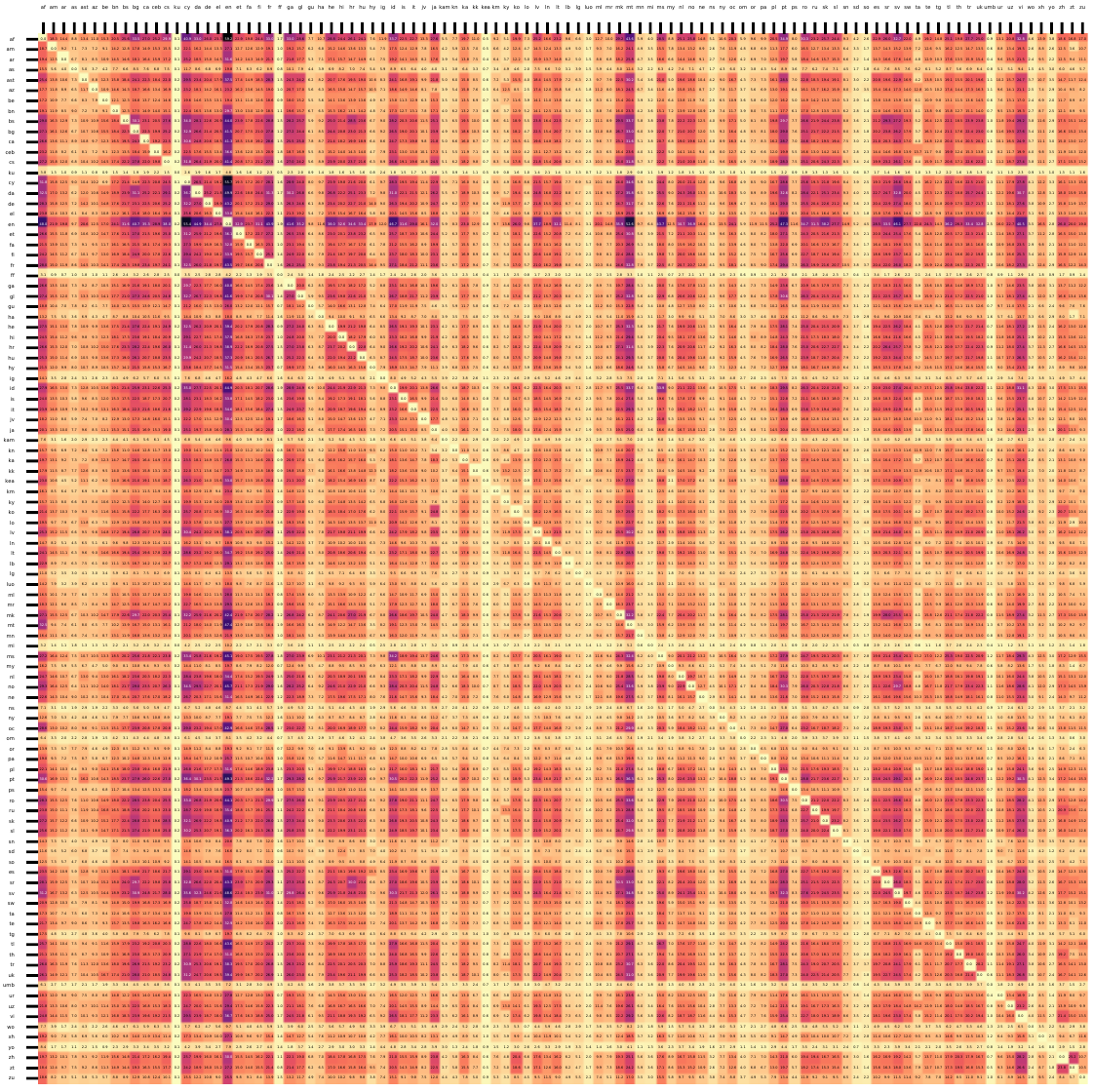}
	\caption{Evaluation results of our multilingual model (24 encoder layers and 6 decoder layers) on all translation directions on the FLORES-101 devtest set. The language $x$ in the $i$-th row and language $y$ in the $j$-th column denotes the translation direction from the language $x$ to language $y$. For example, the cell of the $1$-th row (af) and the $2$-th column (am) represents the result of the translation direction af$\to$am. The table shows the results of all translation directions of 102 languages.}
	\label{full_track_24L_6L}
\end{center}
\end{figure*}

\begin{figure*}[t]
\begin{center}
	\includegraphics[width=1.0\textwidth]{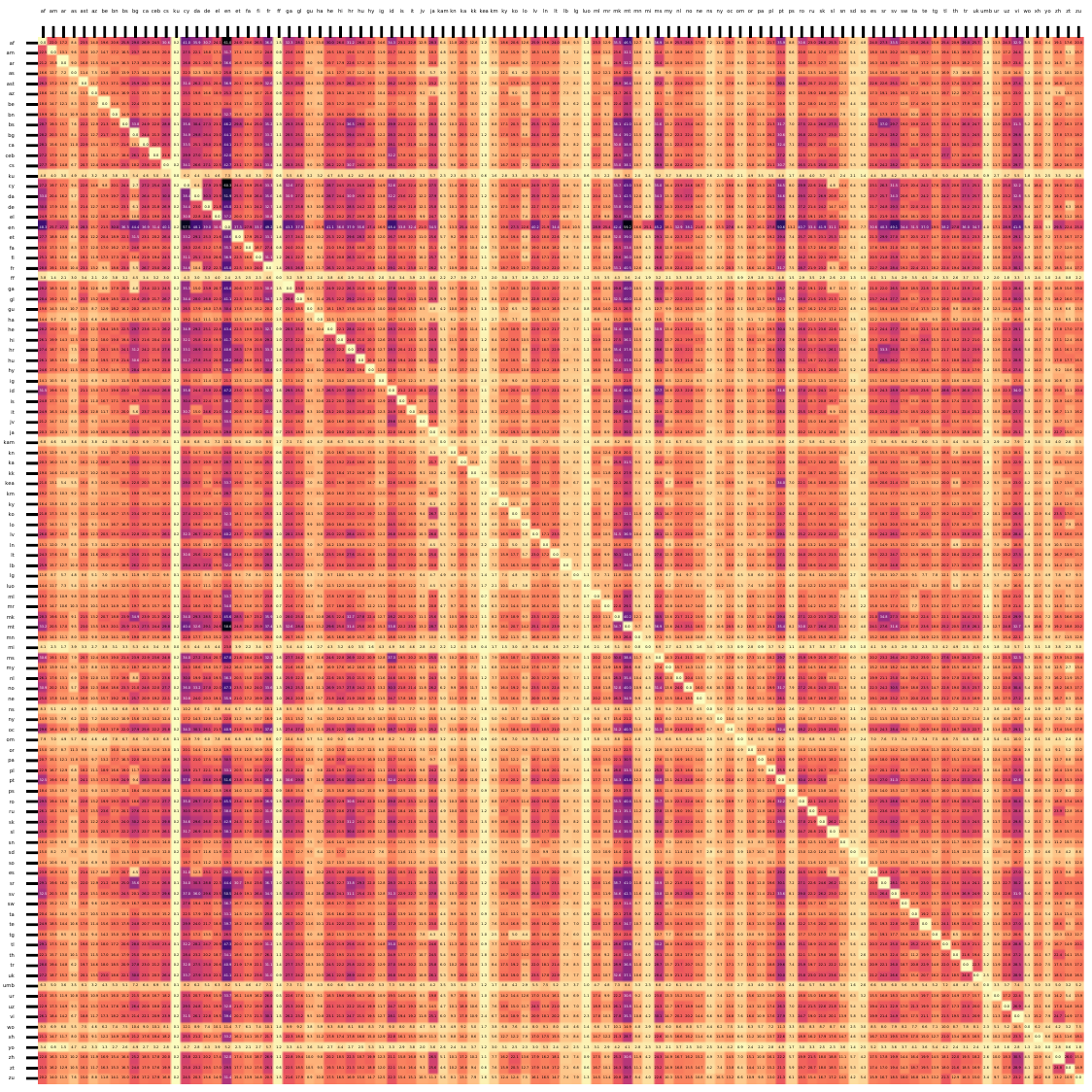}
	\caption{Evaluation results of our multilingual model (36 encoder layers and 12 decoder layers) on all translation directions on the FLORES-101 devtest set. The language $x$ in the $i$-th row and language $y$ in the $j$-th column denotes the translation direction from the language $x$ to language $y$. For example, the cell of the $1$-th row (af) and the $2$-th column (am) represents the result of the translation direction af$\to$am. The table shows the results of all translation directions of 102 languages.}
	\label{full_track_36L_12L}
\end{center}
\end{figure*}

Given the unbalanced large-scale multilingual corpora, we use the hybrid strategy for the translation for Large Track. The pivot-based translation is more suitable for the low-resource translation direction between non-English languages since the corpora of X$\to$Y are commonly scarce. Our model with the 36 encoder layers significantly outperforms the shallow counterpart with the 24 encoder layers, which indicates that using a deep encoder and shallow decoder is a good trade-off between the translation quality and the decoding speed.
Table~\ref{table:large_track} shows that our model with the hybrid strategy gets the best performance with less inference cost than the pivot-based translation which costs double inference time compared to the direct translation. We build a massively multilingual neural machine model, which translates between any pair of 102 languages. In Figure~\ref{full_track_24L_6L} and Figure~\ref{full_track_36L_12L}, we reported the spBLEU scores of the shallow model with 24 encoder layers and 6 decoder layers and our best multilingual model with 36 encoder layers and 12 decoder layers in all translation directions, where the languages are ordered alphabetically by the language code. Nearly 30\% translation directions adopt the pivot-based translation, where the zero-resource and low-resource translation directions lack of supervised training data tend to be chosen for pivot-based translation.

\subsection{Small Track \#1} In Table \ref{table:small_track1}, we compare the performance of M2M with our method in different architectures on any to English (X$\to$En), English to any (En$\to$X), and the translation between any non-English languages (X$\to$Y). Both En$\to$X and X$\to$En contain 5 directions, while X$\to$Y have 20 directions. Given the enormous bilingual and back-translation data of Small Track \#1, we are able to perform the direct translation for all X$\to$Y directions. Furthermore, we explore the deep encoder (36 encoder layers) and shallow decoder (12 decoder layers) considering the limited inference time. From Table~\ref{table:small_track1}, we can observe that the largest model (36 encoder layers and 12 decoder layers) has a significant improvement of +9.41 BLEU points over the strong M2M baseline.

\subsection{Small Track \#2} Compared to Small Track \#1 (273M bilingual pairs), Small Track \#2 contains smaller while more unbalanced training data (93M bilingual pairs). Therefore, we consider the hybrid strategy for the X$\to$Y translation directions. We separately calculate the BLEU scores of the direct and the pivot-based translation on the validation set. For those directions satisfying $\text{BLEU}_{direct}(X, Y) \geq \text{BLEU}_{pivot}(X, Y)$, we employ the direct translation. Otherwise, we use the pivot-based translation direction by first translating the source language to English and then to the target language. According to Table~\ref{table:small_track2}, \textbf{\ourmethod{} (Hybrid)} outperforms both the direct and pivot-based translation by about +0.5 BLEU points. It confirms that the hybrid strategy is essential since the training data of the X$\to$En and En$\to$Y is easy to obtain while the X$\to$Y is hard to obtain. 
The deep model with the 36 encoder layers and 12 decoder layers has comparable performance with the shallow model with the 24 encoder layers and 12 decoder layers, which may be caused by the overfitting problem on the low-resource directions.  

\subsection{Discussion on Progressive Learning}
Given the pre-trained model and large-scale parallel data, we adopt progressive learning as an alternative to fine-tune the multilingual model on the multilingual translation task. 
Our multilingual model is first trained on the large-scale noisy data and then continues to be tuned on the clean data (Noisy Data $\to$ Clean Data), where the model is denoted as {\large \ding{174}}. Since the training data of $K$ languages in the dual-pseudo parallel data is generated by the same English monolingual data, we are able to adopt all possible $K \times (K-1)$ training directions on the clean data. The performance of many translation directions is improved by the additional dual-pseudo data while the performance of other directions has been degraded compared to the initial model {\large{\ding{175}}}, due to the poor quality of some languages in the dual-pseudo data. Therefore, only the part of $K \times (K-1)$ training directions is selected to continue training the multilingual model (Numerous Directions $\to$ Selected Directions), which we denoted as {\large{\ding{173}}}. To further enlarge the model's capacity, we extend the shallow model with 24 encoder layers to the deep model with 36 encoder layers, where the top 12 encoder layers are initialized by random parameters (Shallow Model $\to$ Deep Model). Putting them all together, we obtain the final model {\large{\ding{172}}} \textbf{\ourmethod{}}.
Table~\ref{progressive_learning} summarizes the results of the ablation study of these approaches. It shows that each approach has a significant contribution to the final model. This proves the effectiveness of progressive learning that can gradually improve performance in different aspects.

%%%%%%%%%%%%%%%%%%%%%%%%%%%%%%%%%%%%%%%%%%%%%%%%%%%%%%%%%%%%%%%%%%%%%%%%%%%%
\begin{table}[t]
\centering
\resizebox{1.0\columnwidth}{!}{
\begin{tabular}{c|l|c}
\toprule
ID & Large Track          &  Avg$_{all}$   \\
\midrule
{\large{\ding{172}}} & \ourmethod{}                             &     14.65      \\
{\large{\ding{173}}} & {\large{\ding{172}}} - Shallow Model $\to$ Deep Model               &  13.89         \\
{\large{\ding{174}}} & {\large{\ding{173}}} - Numerous Directions $\to$ Selected Directions     &  13.09             \\
{\large{\ding{175}}} & {\large{\ding{174}}} - Noisy Data $\to$ Clean Data                  &  12.24           \\
\bottomrule
\end{tabular}
}
\caption{Ablation study of the large track on devtest. \ourmethod{} is fine-tuned on the multilingual translation task via progressive learning.}
\label{progressive_learning}
%\vspace{-10pt}
\end{table}
%%%%%%%%%%%%%%%%%%%%%%%%%%%%%%%%%%%%%%%%%%%%%%%%%%%%%%%%%%%%%%%%%%%%%%%%%%%%

\section{Submissions}

Considering the trade-off between the decoding time and the performance, we submit the model (24 encoder layers and 12 decoder layers) with the hybrid strategy to both the Large Track and Small Track \#2, while the deep model (36 encoder layers and 12 decoder layers) with the direct translation is submitted to Small Track \#1. 
Table~\ref{submission} summarizes the evaluation results of our model on the hidden test sets. According to the final results on the leaderboard, \textbf{\ourmethod{}} ranks first across three tracks.

%%%%%%%%%%%%%%%%%%%%%%%%%%%%%%%%%%%%%%%%%%%%%%%%%%%%%%%%%%%%%%%%%%%%%%%%%%%%
\begin{table}[t]
\centering
\resizebox{1.0\columnwidth}{!}{
\begin{tabular}{l|l|c}
\toprule
Track                     &Submission Name &  Avg$_{all}$   \\
\midrule
Large          &\ourmethod{} (Microsoft)            & 16.63          \\
Small \#1      &\ourmethod{} (Microsoft-Small)	  & 37.59            \\
Small \#2      &\ourmethod{} (Microsoft-Small)      & 33.89          \\
\bottomrule
\end{tabular}
}
\caption{Submission results based on the hidden test sets of our method on three tracks, including Large Track, Small Track \#1, and Small Track \#2.}
\label{submission}
% \vspace{-10pt}
\end{table}
%%%%%%%%%%%%%%%%%%%%%%%%%%%%%%%%%%%%%%%%%%%%%%%%%%%%%%%%%%%%%%%%%%%%%%%%%%%%

\section{Conclusion}
This paper describes Microsoft’s submission to the large-scale multilingual machine translation of the WMT21 shared task. Our multilingual translation model achieves substantial improvement over the baseline systems by fine-tuning the pre-trained language model DeltaLM. We further enhance the model performance with the progressive learning and the iterative back-translation methods. As a result, our submitted systems get the top evaluation results on three tracks, including Large Track, Small Track \#1, and Small Track \#2.

% \subsection{Example Study} 
% To inform the model translation direction, the M2M model adds the source language symbol to the source sentence and the target symbol to the beginning of the target sentence. To encourage better cross-lingual representations, we do not use the source language symbol for the inference by only prepending the target language symbol to the source sentence. In Figure \ref{example}, the language token \texttt{[2am]} is prefixed to the source sentence (af) to indicate the am$\to$af direction.

% % \subsection{Analysis}
% % \paragraph{Representation Visualization}

% %%%%%%%%%%%%%%%%%%%%%%%%%%%%%%%%%%%%%%%%%%%%%%%%%%%%%%%
% \begin{figure}[t]
% \centering
% \includegraphics[width=1.0\linewidth]{graph/example.pdf}
% \caption{\label{example} An example from the FLORES-101 af$\rightarrow$am devtest set. Our multilingual model can translate between any pair of 102 languages by prepending the target language symbol to the source sentence. }
% \end{figure}
% %%%%%%%%%%%%%%%%%%%%%%%%%%%%%%%%%%%%%%%%%%%%%%%%%%%%%%%

\bibliography{custom}
\bibliographystyle{acl_natbib}

% \appendix
% \section{Detailed Results of Large Track}
% In Figure 8, We reported the spBLEU scores of our multilingual models with different layers on all language, where the languages are organized alphabetically by language code.
%%%%%%%%%%%%%%%%%%%%%%%%%%%%%%%%%%%%%%%%%%%%%%%%%%%%%%%%%%%
%%%%%%%%%%%%%%%%%%%%%%%%%%%%%%%%%%%%%%%%%%%%%%%%%%%%%%%%%%%

% %%%%%%%%%%%%%%%%%%%%%%%%%%%%%%%%%%%%%%%%%%%%%%%%%%%%%%%%%%%
% \begin{figure*}[t]
% \begin{center}
% 	\includegraphics[width=1.0\textwidth]{graph/24L_12L.pdf}
% 	\caption{Evaluation results of our multilingual model (24 encoder layers and 12 decoder layers) on all translation directions of the FLORES-101 devtest set. The language $x$ in the $i$-th row and language $y$ in the $j$-th denotes the translation direction from the language $x$ to language $y$. For example, the $1$-th row (af) and the $2$-th column (am) represents the result of the translation direction af$\to$am. The table shows the results of all translation directions.}
% 	\label{full_track_24L_12L}
% \end{center}
% \end{figure*}
% %%%%%%%%%%%%%%%%%%%%%%%%%%%%%%%%%%%%%%%%%%%%%%%%%%%%%%%%%%%

%%%%%%%%%%%%%%%%%%%%%%%%%%%%%%%%%%%%%%%%%%%%%%%%%%%%%%%%%%%
%%%%%%%%%%%%%%%%%%%%%%%%%%%%%%%%%%%%%%%%%%%%%%%%%%%%%%%%%%%

% \section{Example Appendix}
% \label{sec:appendix}

%This is an appendix.

\end{document}